\documentclass[11pt]{article}
\usepackage[final]{acl}
\usepackage{times}
\usepackage{latexsym}
\usepackage[T1]{fontenc}
\usepackage[utf8]{inputenc}
\usepackage{microtype}
\usepackage{inconsolata}
\usepackage{graphicx}
\usepackage{subcaption}
\usepackage{amsmath} 
\usepackage{booktabs} 
\usepackage{pgfplots}
\pgfplotsset{compat=1.17}
\title{WebUncertainty: Dual-Level Uncertainty Driven Planning and Reasoning For Autonomous Web Agent}

\author{
  \textbf{Lingfeng Zhang\textsuperscript{1}},
  \textbf{Yongan Sun\textsuperscript{1}},
  \textbf{Jinpeng Hu\textsuperscript{1}},
  \textbf{Hui Ma\textsuperscript{1}},
  \textbf{Ying Yang\textsuperscript{1}},
  \\
  \textbf{Kuien Liu\textsuperscript{1,2}},
  \textbf{Zenglin Shi\textsuperscript{1}\thanks{Corresponding author: zenglin.shi@hfut.edu.cn}},
  \textbf{Meng Wang\textsuperscript{1}}
  \\
  \textsuperscript{1}Hefei University of Technology 
  \quad
  \textsuperscript{2}Academy of Cyber, CETC Group
}
\begin{document}
\maketitle
\begin{abstract}
Recent advancements in large language models (LLMs) have empowered autonomous web agents to execute natural language instructions directly on real-world webpages. However, existing agents often struggle with complex tasks involving dynamic interactions and long-horizon execution due to rigid planning strategies and hallucination-prone reasoning. To address these limitations, we propose WebUncertainty, a novel autonomous agent framework designed to tackle dual-level uncertainty in planning and reasoning. Specifically, we design a Task Uncertainty-Driven Adaptive Planning Mechanism that adaptively selects planning modes to navigate unknown environments. Furthermore, we introduce an Action Uncertainty-Driven Monte Carlo tree search (MCTS) Reasoning Mechanism. This mechanism incorporates the Confidence-induced Action Uncertainty (ConActU) strategy to quantify both aleatoric uncertainty (AU) and epistemic uncertainty (EU), thereby optimizing the search process and guiding robust decision-making. Experimental results on the WebArena and WebVoyager benchmarks demonstrate that WebUncertainty achieves superior performance compared to state-of-the-art baselines.
\end{abstract}

\section{Introduction}

\begin{figure*}[t]
    \centering
    % 第一张图：宽度设为 0.48\textwidth
    \begin{subfigure}[b]{0.48\textwidth} 
        \includegraphics[width=\textwidth]{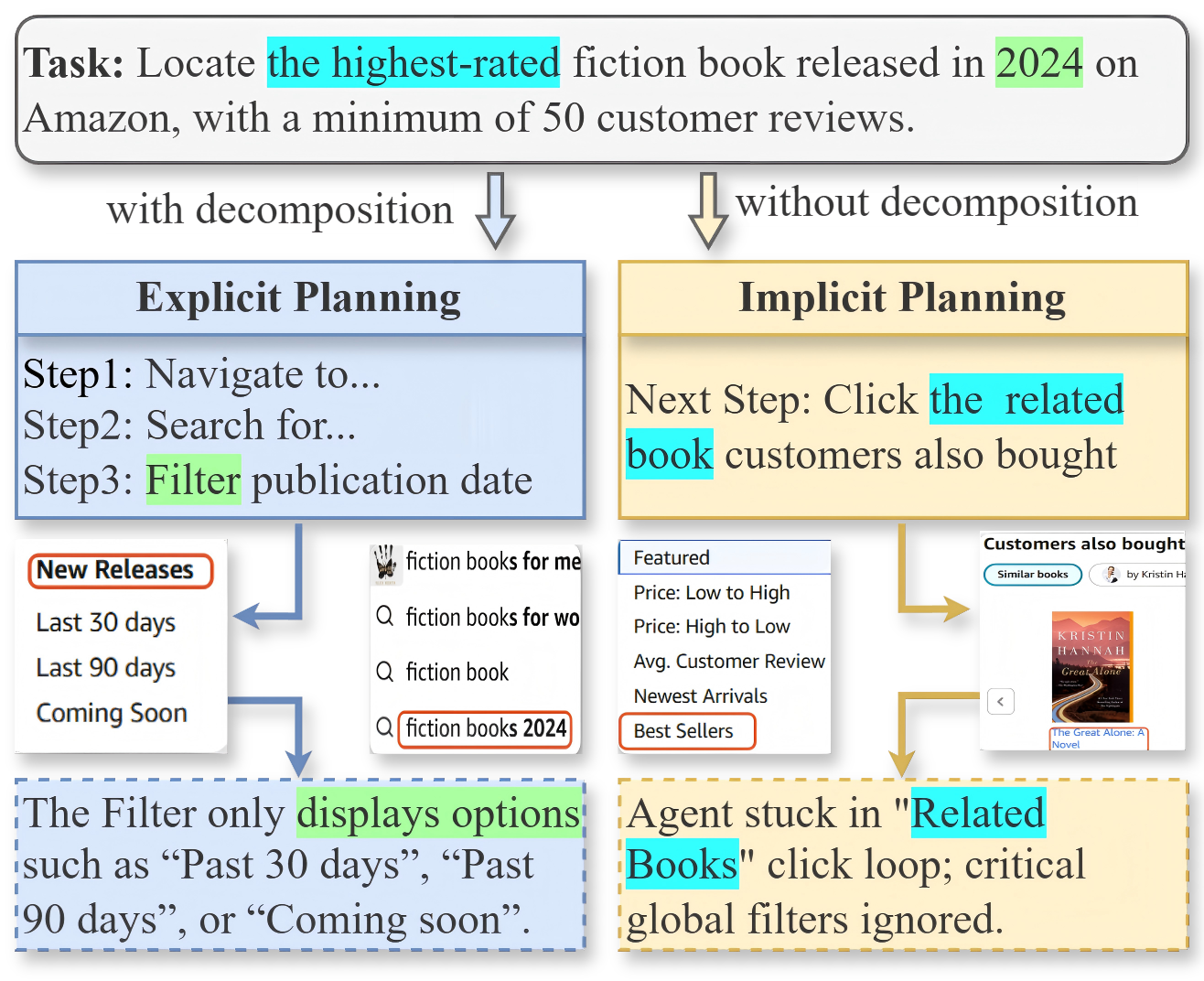}
        \caption{The task uncertainty in planning.}
        \label{fig:challenge_a}
    \end{subfigure}
    \hfill 
    % 第二张图：宽度设为 0.48\textwidth
    \begin{subfigure}[b]{0.48\textwidth} 
        \includegraphics[width=\textwidth]{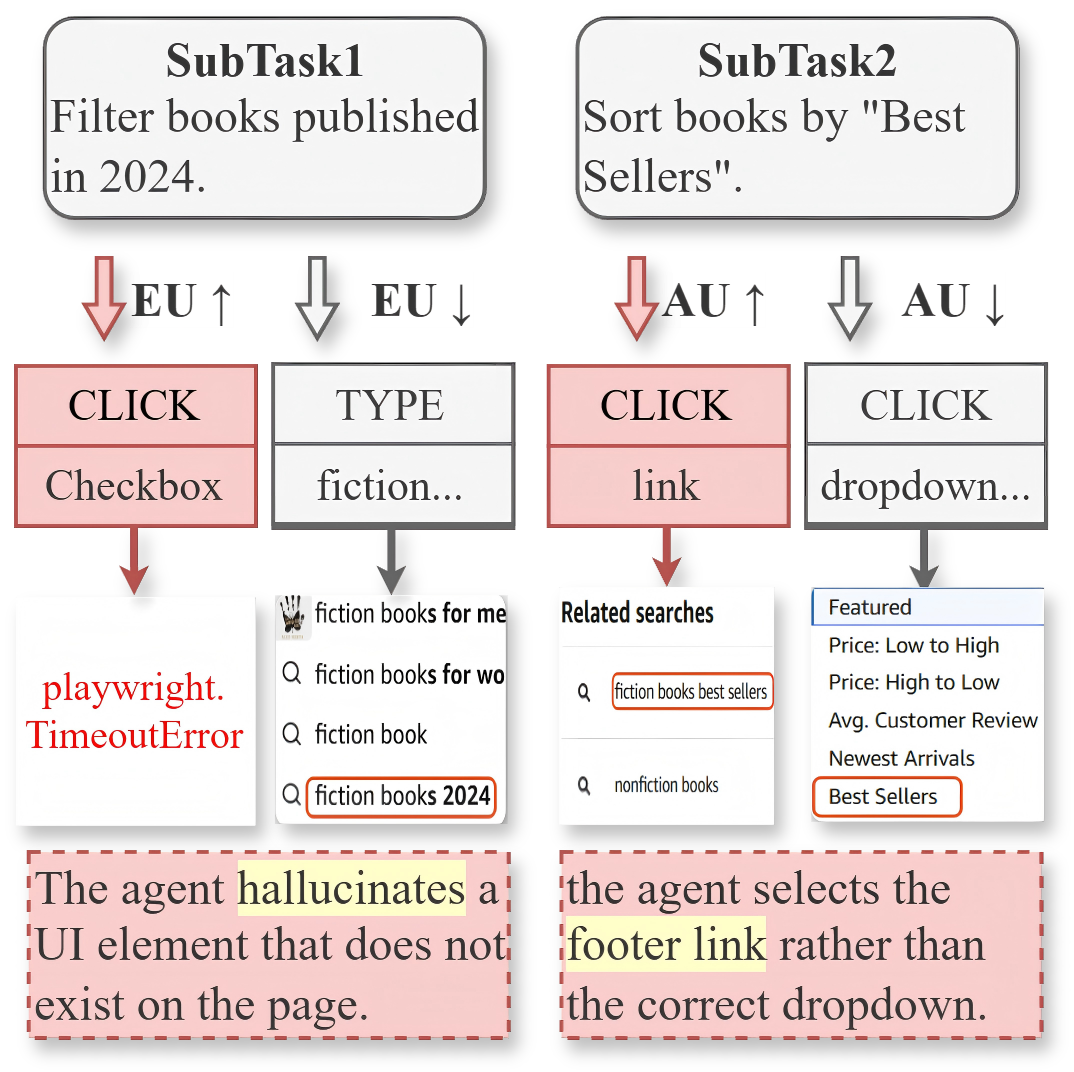}
        \caption{The task uncertainty in reasoning.}
        \label{fig:challenge_b}
    \end{subfigure}
    
    \caption{The dual-level uncertainty challenges in complex web tasks.}
    \label{fig:challenge}
\end{figure*}

Web automation facilitates online services, including information aggregation, transaction processing, and enterprise workflows \citep{deng2023MIND2WEBGeneralistAgent,zheng2024GPT4VisionGeneralistWeb}. However, existing solutions based on hand-crafted scripts, programmatic APIs, and Robotic Process Automation (RPA) tools are brittle and task-specific, often failing under new tasks or minor interface changes\citep{liu2018ReinforcementLearningWeb,pu2023DiLogicsCreatingWeb}. Recent advances in large language models (LLMs) with strong natural language understanding and reasoning capabilities \citep{deng2024LargeLanguageModel,du2026SurveyOptimizationLarge, zhang2026System1System} enable more flexible web agents that execute instructions directly on real-world webpages \citep{hu2025OSAgentsSurvey, nguyen2025GUIAgentsSurvey, ning2025SurveyWebAgentsNextgeneration}. To enhance the reliability of these agents, recent studies have equipped them with planning mechanisms \citep{erdogan2025PLANANDACTImprovingPlanning, luo2025BrowsingHumanMultimodal, shahnovsky2026AIPlanningFramework} to decompose user instructions into manageable subgoals, and reasoning mechanisms \citep{koh2024TreeSearchLanguage, zhang2025WebPilotVersatileAutonomous, wei2026AgenticReasoningLarge} to guide the decision-making process. Despite these advancements, current agents still struggle with complex tasks requiring dynamic interaction and long-horizon execution \citep{wu2025WebWalkerBenchmarkingLLMs, yang2025AgenticWebWeaving}.

First, complex tasks involve dynamic web interactions. This dynamism makes it difficult for preplanned subgoals to adapt to unknown environments \citep{he2024WebVoyagerBuildingEndtoEnd, zhou2024WebArenaRealisticWeb}. As shown in Figure~\ref{fig:challenge_a}, an agent employing one-shot explicit planning intends to use the ``Publication Year'' filter to select 2024. However, it overlooks that the Amazon sidebar lacks this option, resulting in an execution failure. Conversely, iteratively generating subgoals via implicit planning introduces a different risk. It can distract the agent with the highest-rated book on the current page, causing it to neglect the global rating filter and fall into a local optimum. The agent can effectively resolve these issues by flexibly selecting its planning mode based on the webpage state and task progress. For instance, the agent can use implicit planning to correct cognitive biases during date filtering, and explicit planning to reduce local noise during rating sorting \citep{luo2025BrowsingHumanMultimodal}.

Second, complex tasks involve long-horizon execution, where reasoned actions are highly prone to errors due to LLM hallucinations and the snowball effect \citep{gan2025RethinkingExternalSlowthinking, xia2025SurveyUncertaintyEstimation}. As shown in Figure~\ref{fig:challenge_b}, the agent may operate on nonexistent elements due to a lack of domain-specific knowledge, or select incorrect elements due to the probabilistic nature of LLMs. This issue primarily stems from an overreliance on LLM-generated actions without considering their uncertainty \citep{zhang2025WebPilotVersatileAutonomous,zhao2025UncertaintyPropagationLLM}. Recent studies \citep{ma2025EstimatingLLMUncertainty} have introduced logits-induced token uncertainty to decouple LLM uncertainty into aleatoric uncertainty (AU) and epistemic uncertainty (EU). However, these approaches focus primarily on discrete tokens, overlooking the semantic meaning of the generated actions.

In this work, we propose \textbf{WebUncertainty}, an autonomous web agent designed to address complex tasks requiring dynamic interactions and long-horizon execution by tackling the dual-level uncertainty arising from planning and reasoning. At the planning level, we design a Task Uncertainty-Driven Adaptive Planning Mechanism. Prior to each planning step, an analysis agent evaluates the task uncertainty based on the current state and task progress. Subsequently, a planning agent adaptively selects the appropriate planning mode based on this uncertainty to effectively handle unknown environments. At the reasoning level, we design an Action Uncertainty-Driven  Monte Carlo tree search (MCTS) Reasoning Mechanism. During the MCTS expansion phase, a reasoning agent generates multiple candidate actions along with their confidence scores. We introduce the Confidence-induced Action Uncertainty (ConActU) strategy to quantify action uncertainty at both the AU and EU levels. Finally, we optimize the MCTS search process by combining this quantified uncertainty with feedback from an evaluation agent.

Our contributions are summarized as follows:
\begin{itemize}
    \setlength{\itemsep}{0pt}
    \setlength{\parsep}{0pt}
    \item We propose WebUncertainty, a novel autonomous web agent framework that addresses dual-level uncertainty in planning and reasoning, achieving robust performance in complex tasks involving dynamic interactions and long-horizon execution.
    \item We design a Task Uncertainty-Driven Adaptive Planning Mechanism, which adaptively switches planning modes based on dynamic environmental changes, enabling the system to effectively align sub-goals with unpredictable web environments.
    \item We introduce an Action Uncertainty-Driven MCTS Reasoning Mechanism, incorporating the ConActU strategy that quantifies both AU and EU to guide the search process, thereby mitigating hallucinations and ensuring reliable decision-making.
\end{itemize}

Experimental results on WebArena\citep{zhou2024WebArenaRealisticWeb} and WebVoyager\citep{he2024WebVoyagerBuildingEndtoEnd} demonstrate that our WebUncertainty achieves superior performance, particularly for complex tasks, outperforming existing methods.\footnote{Code is available at: \url{https://github.com/windbd/WebUncertainty}}

\section{Related Work}

\paragraph{Web Agents} 
A web agent is an autonomous AI system that perceives web interfaces through Document Object Model (DOM) trees or screenshots, makes decisions, and executes actions to follow natural language instructions \citep{gur2024RealworldWebAgentPlanning,nguyen2025GUIAgentsSurvey, ning2025SurveyWebAgentsNextgeneration}. Early approaches primarily relied on rule-based systems or imitation learning, which required extensive human demonstration and were brittle to interface changes \citep{liu2018ReinforcementLearningWeb, pu2023DiLogicsCreatingWeb}. The emergence of LLMs has revolutionized this field \citep{deng2024LargeLanguageModel}. By leveraging their powerful natural language understanding and generation capabilities, modern agents generalize across diverse websites \citep{song2025BrowsingAPIbasedWeb,lai2025WebGLMEfficientReliable, gupta2026MolmoWebOpenVisual,zhang2026WebNavigatorGlobalWeb}. However, deploying these agents in real-world scenarios remains challenging due to the dynamic nature of web environments and the complexity of long-horizon interactions \citep{huang2025R2D2RememberingReplaying,he2025OpenWebVoyagerBuildingMultimodal}.

\paragraph{Planning Mechanisms in Agents} 
Planning serves as the strategic core of web agents, responsible for decomposing high-level instructions into executable subgoals \citep{zhang2024AskbeforeplanProactiveLanguage, xi2025RisePotentialLarge,shahnovsky2026AIPlanningFramework}. Existing planning strategies are generally categorized into: 1) explicit planning, which involves formal task decomposition \citep{li2023ZeroShotLanguageAgent, niu2024ScreenAgentVisionLanguage, zheng2024GPT4VisionGeneralistWeb}, and 2) implicit planning, where agents predict actions reactively without a formal decomposition phase \citep{koh2024TreeSearchLanguage, he2025OpenWebVoyagerBuildingMultimodal, zhang2025WebPilotVersatileAutonomous}. One-shot explicit planning generates a complete sequence of actions upfront but lacks adaptability; for instance, pregenerated plans quickly become obsolete if the web environment shifts, such as when a pop-up appears. Iterative approaches address this via replanning at fixed steps, yet these methods typically employ rigid protocols without assessing the necessity of such adjustments. Crucially, current approaches fail to model Task Uncertainty \citep{ning2025SurveyWebAgentsNextgeneration}. They do not dynamically adapt their planning mode between explicit and implicit planning based on the agent's familiarity with the environment, leading to either inefficiency in simple tasks or failure in complex, unknown domains \citep{zhou2024WebArenaRealisticWeb, he2024WebVoyagerBuildingEndtoEnd}.

\paragraph{Reasoning Mechanisms in Agents} 
Reasoning serves as the decision-making core of web agents, translating planned subgoals into atomic actions \citep{pahuja2025ExplorerScalingExplorationdriven,wei2026AgenticReasoningLarge, zhang2026System1System}. Existing methods range from reactive reasoning \citep{abuelsaad2024AgentEAutonomousWeb, yang2025AgentOccamSimpleStrong} to strategic reasoning that employs tree search to explore trajectories \citep{koh2024TreeSearchLanguage, yu2025ExACTTeachingAI,zhang2025WebPilotVersatileAutonomous}. Crucially, most reasoning mechanisms overlook the risk of hallucinations, allowing execution errors, such as operating on nonexistent elements, to propagate through long-horizon tasks and lead to cascading failures \citep{gan2025RethinkingExternalSlowthinking,zhao2025UncertaintyPropagationLLM}. While \citet{ma2025EstimatingLLMUncertainty} disentangled AU and EU using logits to identify hallucinations, their approach remains confined to discrete tokens and overlooks action semantics. WebUncertainty addresses this gap by incorporating the ConActU strategy into MCTS, explicitly quantifying these uncertainty dimensions at the action level to ensure semantically grounded decision-making.

\section{Methodology}

\begin{figure*}[t]
  \centering
  \includegraphics[width=\textwidth]{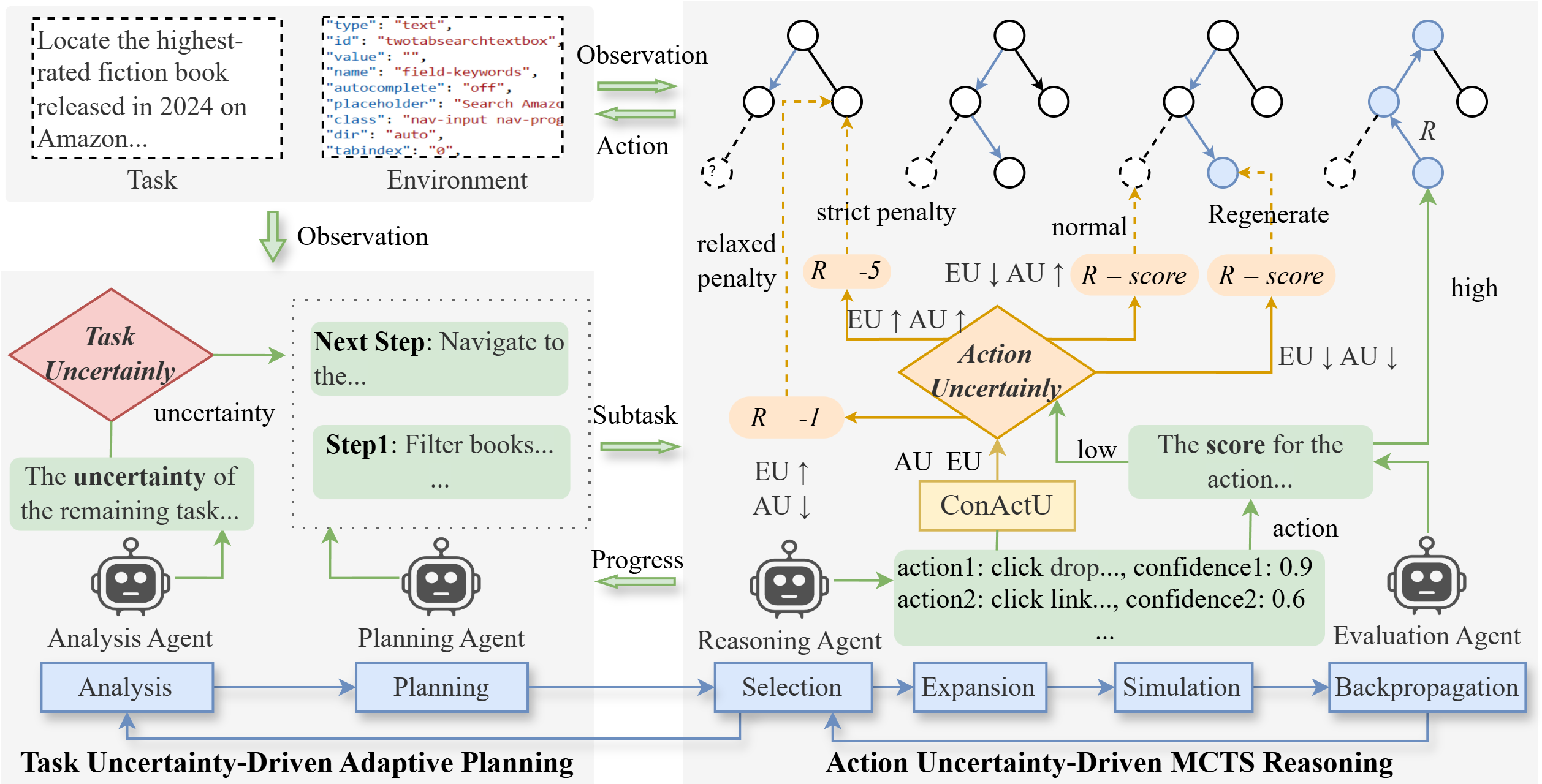}
  \caption{Overview of WebUncertainty. The framework decouples the process into Task Uncertainty-Driven Adaptive Planning (left) and Action Uncertainty-Driven MCTS Reasoning (right).}
  \label{fig:methodology}
\end{figure*}

As illustrated in Figure~\ref{fig:methodology}, we propose WebUncertainty, a hierarchical framework that tackles dual-level uncertainty for web agents. The framework consists of two core components:
(1) A Task Uncertainty-Driven Adaptive Planning Mechanism (Section~\ref{sec:planning}). In this stage, an Analysis Agent evaluates task uncertainty based on the environment and task progress. A Planning Agent then adaptively switches planning modes to ensure that subgoals align with the evolving webpage state.
(2) An Action Uncertainty-Driven MCTS Reasoning Mechanism (Section~\ref{sec:reasoning}). A Reasoning Agent integrates the ConActU strategy to quantify both AU and EU. An Evaluation Agent then assesses action scores to mitigate hallucinations and guide robust execution.

Formally, we model the web navigation task as a Partially Observable Markov Decision Process (POMDP). Given a global instruction $I$ and a webpage observation $O_t$, the agent operates hierarchically to generate an atomic action $a_t = (e, o, v)$ at each step. Here, $e$ denotes the interactive element, $o$ specifies the operation type (e.g., click or type), and $v$ represents the optional value. The objective is to generate an optimal action sequence that maximizes the success probability of the instruction $I$.

\subsection{Task Uncertainty-Driven Adaptive Planning Mechanism}
\label{sec:planning}

Static planning strategies often fail in complex web tasks. Explicit decomposition struggles with unknown environments, while implicit stepping risks falling into local optima. To address these issues, we propose the Task Uncertainty-Driven Adaptive Planning Mechanism. Before each planning step, an Analysis Agent evaluates task uncertainty based on the current webpage state and task progress. A Planning Agent then dynamically selects the optimal planning mode. It leverages implicit planning to adapt to unpredicted changes when uncertainty is high. Conversely, it employs explicit planning to maintain global coherence when uncertainty is low. This adaptive approach ensures that generated subgoals effectively align with the evolving web environment.

\paragraph{Task Uncertainty Analysis}
At each time step $t$, an Analysis Agent ($\pi_{\text{ana}}$) evaluates the task before plan generation. It processes the global instruction $I$, the current webpage observation $O_t$, and the execution history $H_t$. Its primary objective is to determine the remaining task objectives $T_{\text{rem}}$ and quantify the associated task uncertainty $u_{\text{plan}} \in [0, 1]$, formulated as:
\begin{equation}
    T_{\text{rem}}, u_{\text{plan}} = \pi_{\text{ana}}(I, O_t, H_t) \label{eq:uncertainty_analysis}
\end{equation}
Here, $T_{\text{rem}}$ represents the pending goals. The scalar $u_{\text{plan}}$ links environmental unfamiliarity to execution complexity. A high $u_{\text{plan}}$ indicates an unfamiliar webpage structure where achieving $T_{\text{rem}}$ is highly complex. Conversely, a low $u_{\text{plan}}$ suggests a familiar environment with minimal task complexity.

\paragraph{Adaptive Task Planning}
Guided by the task uncertainty $u_{\text{plan}}$, the Planning Agent ($\pi_{\text{plan}}$) selects a subgoal generation strategy based on a threshold $\delta$. In low-uncertainty scenarios ($u_{\text{plan}} \le \delta$), the agent activates the Explicit Planner ($\pi_{\text{plan}}^{\text{exp}}$) to perform one-shot decomposition. It then commits to the first subgoal in the generated sequence, formulated as $g_t = \text{First}(\pi_{\text{plan}}^{\text{exp}}(T_{\text{rem}}, O_t))$, to ensure long-horizon coherence. 

Conversely, in high-uncertainty scenarios ($u_{\text{plan}} > \delta$), the agent shifts to the Implicit Planner ($\pi_{\text{plan}}^{\text{imp}}$) to adapt flexibly to unpredicted environmental dynamics. In this mode, the agent directly predicts the immediate subgoal as $g_t = \pi_{\text{plan}}^{\text{imp}}(T_{\text{rem}}, O_t)$. The resulting subgoal $g_t$ then directs the subsequent reasoning phase. As execution proceeds, updated observations and task progress may reduce uncertainty, enabling a dynamic switch from implicit exploration back to explicit execution.

\subsection{Action Uncertainty-Driven MCTS Reasoning Mechanism}
\label{sec:reasoning}

After planning, the agent resolves the atomic subgoal $g_t$ during the execution phase. We model this process as a tree search where nodes represent webpage states and edges denote concrete actions. To navigate vast action spaces and mitigate hallucinations, we propose the Action Uncertainty-Driven MCTS Reasoning Mechanism. This module employs the ConActU strategy to guide the four phases of MCTS:

\paragraph{Selection}
The agent traverses the tree from the root. At each step, it selects the child node that maximizes the predictor-corrected upper confidence bound (PUCT). We integrate the action confidence from the ConActU strategy as a prior to guide the search:
\begin{equation}
    a_t = \operatorname*{argmax}_{a \in \mathcal{A}} \left[ Q(s, a) +  U(s, a) \right] \label{eq:selection_argmax}
\end{equation}
\begin{equation}
    U(s, a) = w_{\text{puct}} \cdot \frac{P_{\text{con}}(s, a) \sqrt{\sum_{b} N(s, b)}}{1 + N(s, a)} \label{eq:puct_bonus}
\end{equation}
Here, $Q(s, a)$ is the value estimate and $N(s, a)$ is the visit count. $P_{\text{con}}(s, a)$ represents the confidence score computed during expansion. This mechanism ensures that the search prioritizes actions with higher evidential support.

\paragraph{Expansion}
Upon reaching a leaf node, the reasoning agent generates $K$ candidate actions and directly outputs their corresponding confidence scores $\mathbf{c} = [c_1, c_2, \dots, c_K]$. To quantify uncertainty, we employ the ConActU strategy. First, we normalize the scores into a pseudo-probability distribution $p_i = c_i / \sum_{j=1}^K c_j$. We then compute the average confidence as a total evidence proxy $E = \frac{1}{K} \sum_{i=1}^K c_i$. To measure the competition among candidates, we calculate the normalized predictive entropy $H_{\text{norm}} = -\frac{1}{\log K} \sum_{i=1}^K p_i \log p_i$. Based on these metrics, we formulate EU and AU as follows:
\begin{equation}
    \text{EU} = 1 - E \label{eq:eu}
\end{equation}
\begin{equation}
    \text{AU} = H_{\text{norm}} \cdot E \label{eq:au}
\end{equation}
In this formulation, EU captures the hallucination risk derived from a lack of overall confidence. Conversely, AU isolates the inherent ambiguity that occurs when the model possesses knowledge (high $E$) but faces competing valid options (high $H_{\text{norm}}$). Finally, all candidate actions are added to the search tree with their prior probability set to $P_{\text{con}} = p_i$.

\paragraph{Simulation}
Instead of random rollouts, an evaluation agent assesses the potential of the new state to yield a base feasibility score $S_{\text{base}}$. If the score indicates success ($S_{\text{base}} \ge \tau$), the action is accepted, and we directly assign the reward $R = S_{\text{base}}$. For low scores ($S_{\text{base}} < \tau$), we employ an uncertainty-aware modulation strategy to process the failure. The handling method and exploratory purpose for each condition are defined as follows:
\begin{enumerate}
    \item High EU and High AU (Strict Penalty): The state is chaotic and unreliable. We assign a severe penalty ($R = -5$) to strictly prohibit the search from selecting this path in the future.
    \item High EU and Low AU (Relaxed Penalty): The agent lacks domain knowledge, implying a hallucination. We assign a standard penalty ($R = -1$) to encourage the search to backtrack and explore the parent's sibling nodes.
    \item Low EU and High AU (Normal): The agent possesses knowledge but faces stochastic ambiguity. We retain the base score ($R = S_{\text{base}}$) to encourage the search to select different candidate actions under the same node.
    \item Low EU and Low AU (Regenerate): The agent is confident, but the execution yields a low score. This indicates a deterministic error. We assign a zero reward ($R = 0$) to trigger the agent to regenerate new actions based on the current node.
\end{enumerate}

\paragraph{Backpropagation}
Finally, the modulated reward $R$ is backpropagated to update the statistics of all ancestor nodes along the trajectory. We employ an iterative mean update rule to ensure value stability:
\begin{equation}
    N(s, a) \leftarrow N(s, a) + 1 \label{eq:visit_update}
\end{equation}
\begin{equation}
    Q(s, a) \leftarrow Q(s, a) + \frac{R - Q(s, a)}{N(s, a)} \label{eq:value_update}
\end{equation}
This uncertainty-aware update ensures the MCTS converges to a robust policy that avoids epistemic ignorance while managing aleatoric ambiguity.

\section{Experiments}

\subsection{Experimental Setup}

\paragraph{Datasets}
We evaluate WebUncertainty on two benchmarks designed for complex, long-horizon web tasks. WebArena \citep{zhou2024WebArenaRealisticWeb} serves as the primary simulation environment. It comprises 812 tasks derived from realistic platforms, such as GitLab and Reddit. Following \citet{zhang2025WebPilotVersatileAutonomous}, we adopt the text-only setting based on accessibility trees to focus on semantic reasoning. For live web evaluation, we utilize WebVoyager \citep{he2024WebVoyagerBuildingEndtoEnd}. To ensure reproducibility and objectivity, we employ a curated subset of 129 tasks across 13 diverse environments, including Amazon and Google Maps. We strictly exclude unstable pages and open-ended questions to focus on deterministic outcomes.

\paragraph{Metrics}
We report \textbf{Success Rate} (SR) as the primary metric for functional correctness across all experiments.

\paragraph{Compared Baselines}
To evaluate WebUncertainty, we compare it against four state-of-the-art agents representing distinct paradigms. Browser Use\footnote{\url{https://github.com/browser-use/browser-use}} serves as a fundamental baseline for standard web automation. Agent-E \citep{abuelsaad2024AgentEAutonomousWeb} benchmarks our task uncertainty-driven planning against conventional hierarchical architectures. WebPilot \citep{zhang2025WebPilotVersatileAutonomous} utilizes MCTS, providing a direct comparison for our action uncertainty-driven strategy. Finally, AgentOccam \citep{yang2025AgentOccamSimpleStrong} evaluates the agent's robustness in observation-action alignment.

\paragraph{Implementation Details}
To ensure a fair comparison and assess generalizability, we conduct all experiments using two distinct LLM backbones: Qwen-Max-2025-01-25 and GPT-4-turbo-2024-04-09. We execute WebUncertainty and all baselines independently on each backbone. This setup disentangles architectural contributions from the underlying model capabilities. For both LLMs, we fix the temperature at 0.3. In the MCTS reasoning module, we set the maximum node expansion limit to 10 per subgoal and the exploration weight $w_{\text{puct}}$ to 5. These settings balance exploration breadth with exploitation efficiency.

\subsection{Results on WebArena}
\label{sec:results_webarena}

\begin{table*}[t]
\centering
\resizebox{\textwidth}{!}{%
\begin{tabular}{llccccccc}
\toprule
\textbf{Agent} & \textbf{Backbone} & \textbf{SR (\%)} & \textbf{Shop} & \textbf{Admin} & \textbf{GitLab} & \textbf{Map} & \textbf{Reddit} & \textbf{Multi} \\
\midrule
WebArena-rep & GPT-4-Turbo & 16.5 & 16.6 & 15.9 & 10.0 & 22.9 & 21.7 & 16.7 \\
Browser Use & GPT-4-Turbo & 16.9 & 15.0 & 17.6 & 10.6 & 23.9 & 23.6 & 14.6 \\
Agent-E & GPT-4-Turbo & 13.9 & 13.4 & 10.4 & 11.1 & 19.3 & 20.8 & 12.5 \\
WebPilot & GPT-4-Turbo & 37.6 & 41.2 & 43.4 & 33.3 & 37.6 & 37.7 & 16.7 \\
AgentOccam & GPT-4-Turbo & 43.1 & 40.6 & 45.6 & 37.8 & \textbf{46.8} & 61.3 & 14.6 \\
\textbf{WebUncertainty} & GPT-4-Turbo & \textbf{46.9} & \textbf{47.6} & \textbf{49.5} & \textbf{40.0} & 45.9 & \textbf{67.0} & \textbf{18.8} \\
\midrule
Agent-E & Qwen-Max & 14.2 & 12.3 & 9.9 & 12.8 & 17.4 & 23.6 & 14.6 \\
WebPilot & Qwen-Max & 34.5 & 40.1 & 33.0 & 29.4 & \textbf{40.4} & 36.8 & \textbf{18.8} \\
AgentOccam & Qwen-Max & 38.4 & 41.2 & \textbf{42.9} & 33.3 & 27.5 & 55.7 & 16.7 \\
\textbf{WebUncertainty} & Qwen-Max & \textbf{40.1} & \textbf{42.8} & 38.5 & \textbf{37.8} & 38.5 & \textbf{57.5} & 10.4 \\
\bottomrule
\end{tabular}%
}
\caption{Performance comparison on WebArena. The SR is reported over 812 tasks across domains: Shopping, Shopping Admin, GitLab, Map, Reddit, and Multisite. The best results for each backbone group are highlighted in \textbf{bold}.}
\label{tab:webarena_results}
\end{table*}

Table~\ref{tab:webarena_results} presents the comparative analysis on the WebArena benchmark. WebUncertainty establishes a new state-of-the-art. It achieves an overall SR of 46.9\% with GPT-4-Turbo. This performance surpasses the strong baseline AgentOccam (43.1\%) and significantly outperforms the search-based competitor WebPilot (37.6\%). These results empirically validate our dual-level uncertainty framework. It effectively mitigates the rigid planning and hallucination issues that hinder conventional agents in complex, long-horizon tasks.

\paragraph{Adaptability in High-Uncertainty Domains}
Disaggregated analysis reveals that WebUncertainty excels in domains with high ambiguity and interaction complexity. The Reddit domain involves dense textual content and ambiguous user intents. Here, our agent achieves a 67.0\% SR. It surpasses AgentOccam (61.3\%) and nearly doubles WebPilot's performance (37.7\%). This gain is attributed to the Action Uncertainty-Driven MCTS Reasoning Mechanism. By quantifying AU, the agent identifies ambiguous states with multiple plausible actions (High AU). It then prioritizes exploration over premature commitment to avoid local optima.

The GitLab domain requires precise execution of long-horizon workflows. In this domain, our method achieves a 40.0\% SR, compared to WebPilot's 33.3\%. This improvement validates the Task Uncertainty-Driven Adaptive Planning Mechanism. The agent dynamically switches between explicit decomposition for global coherence and implicit stepping for unexpected environmental states. This ensures robust navigation in technical environments.

\paragraph{Robustness Across Reasoning Backbones}
To assess architectural generalizability, we evaluate performance using Qwen-Max. As shown in the bottom section of Table~\ref{tab:webarena_results}, WebUncertainty maintains its lead with an overall SR of 40.1\%. It outperforms AgentOccam (38.4\%) and WebPilot (34.5\%).

Notably, our framework powered by Qwen-Max outperforms the GPT-4-Turbo version of WebPilot (37.6\%). This result underscores the efficacy of the ConActU strategy. By explicitly quantifying EU, our framework enables weaker models to detect their own knowledge boundaries. They can then prune hallucinated actions (High EU) before execution. This uncertainty-aware filtering effectively compensates for the lower intrinsic reasoning capability of the backbone model. It prevents the snowball effect of errors common in long-horizon tasks.

\subsection{Results on WebVoyager}
\label{sec:results_webvoyager}

\begin{table}[t]
\centering
\resizebox{0.48\textwidth}{!}{%
\begin{tabular}{llc}
\toprule
\textbf{Agent} & \textbf{Backbone} & \textbf{SR (\%)} \\
\midrule
WebVoyager-rep & GPT-4-Turbo & 50.4 \\
Browser Use & GPT-4-Turbo & 51.9 \\
Agent-E & GPT-4-Turbo & 59.7 \\
WebPilot & GPT-4-Turbo & 62.0 \\
AgentOccam & GPT-4-Turbo & 64.3 \\
\textbf{WebUncertainty} & GPT-4-Turbo & \textbf{65.9} \\
\midrule
WebVoyager-rep & Qwen-Max & 46.5 \\
Browser Use & Qwen-Max & 48.8 \\
Agent-E & Qwen-Max & 54.3 \\
WebPilot & Qwen-Max & 55.8 \\
AgentOccam & Qwen-Max & 58.9 \\
\textbf{WebUncertainty} & Qwen-Max & \textbf{63.6} \\
\bottomrule
\end{tabular}%
}
\caption{SR comparison on the WebVoyager benchmark. The evaluation is conducted on a curated subset of 129 tasks involving real-world websites with deterministic outcomes. Best results for each backbone are highlighted in \textbf{bold}.}
\label{tab:webvoyager_results}
\end{table}

We extend our evaluation to WebVoyager to assess robustness in live, open-domain web environments. Unlike the controlled simulation of WebArena, WebVoyager involves real-world websites, such as Amazon and Google Maps. These sites feature dynamic content loading, complex DOM structures, and potential network latency.

\paragraph{Robustness in Dynamic Real-World Settings}
As detailed in Table~\ref{tab:webvoyager_results}, WebUncertainty consistently achieves the highest SR across both backbone models. With GPT-4-Turbo, our method attains a 65.9\% SR. It outperforms the strongest baseline AgentOccam (64.3\%) and the search-based WebPilot (62.0\%). AgentOccam enhances performance by optimizing observation grounding. However, it often struggles to recover from execution failures caused by unpredicted interface changes, such as pop-ups or layout shifts. Our framework addresses this limitation through the Action Uncertainty-Driven MCTS Reasoning Mechanism. The ConActU strategy distinguishes between epistemic hallucinations and aleatoric environmental noise. It penalizes high-risk paths and encourages the exploration of alternative actions during confident but unsuccessful executions (Low EU, High AU).

\paragraph{Efficiency on Weaker Backbones}
Results on the Qwen-Max backbone demonstrate the architectural efficiency of our approach. WebUncertainty achieves a 63.6\% SR, outperforming AgentOccam (58.9\%) and WebPilot (55.8\%) by a substantial margin. Notably, our framework powered by the weaker Qwen-Max model outperforms the GPT-4-Turbo version of WebPilot (63.6\% vs. 62.0\%). This highlights a critical insight. In complex web navigation, raw LLM reasoning capability faces diminishing returns without effective uncertainty management. Our framework models Task Uncertainty to adaptively switch planning modes. It also uses Action Uncertainty to prune search trees. This dual-level strategy empowers weaker models to achieve performance levels comparable to, or exceeding, stronger models that rely on standard architectures.

\subsection{Ablation Studies}
\label{sec:ablation}

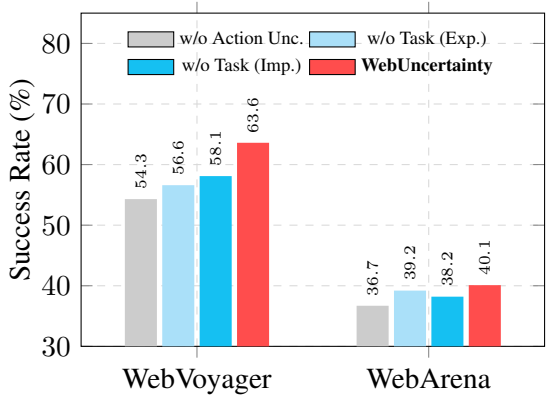
\begin{figure}[t]
    \centering
    \begin{tikzpicture}
        \begin{axis}[
            ybar,
            area legend,          
            bar width=12pt,
            width=0.48\textwidth,
            height=6cm,
            enlarge x limits=0.5,
            symbolic x coords={WebVoyager, WebArena}, 
            xtick=data,
            ymin=30, ymax=85, 
            ylabel={Success Rate (\%)},
            ylabel style={yshift=-5pt, align=center}, 
            legend style={
                at={(0.5,0.98)},
                anchor=north,    
                legend columns=2, 
                draw=none,       
                fill=none,       
                font=\scriptsize 
            },
            nodes near coords, 
            every node near coord/.append style={font=\tiny, rotate=90, anchor=west}, 
            grid=major,
            grid style={dashed, gray!30},
            ytick={30,40,50,60,70,80} 
        ]
        
        % 1. w/o Action Unc.
        \addplot[fill=gray!40, draw=none] coordinates {
            (WebVoyager, 54.3) (WebArena, 36.7)
        };
        
        % 2. w/o Task Unc. (Explicit)
        \addplot[fill=cyan!30, draw=none] coordinates {
            (WebVoyager, 56.6) (WebArena, 39.2)
        };

        % 3. w/o Task Unc. (Implicit)
        \addplot[fill=cyan!70, draw=none] coordinates {
            (WebVoyager, 58.1) (WebArena, 38.2)
        };

        % 4. WebUncertainty (Ours)
        \addplot[fill=red!70, draw=none] coordinates {
            (WebVoyager, 63.6) (WebArena, 40.1)
        };

        \legend{w/o Action Unc., w/o Task (Exp.), w/o Task (Imp.), \textbf{WebUncertainty}}
        \end{axis}
    \end{tikzpicture}
    \caption{Ablation study using the Qwen-Max backbone.}
    \label{fig:ablation}
\end{figure}

To disentangle the contributions of individual components within our framework, we conduct ablation studies on both WebVoyager and WebArena benchmarks using the Qwen-Max backbone. We introduce three variants to strictly isolate the efficacy of the Task Uncertainty-Driven Adaptive Planning Mechanism and the Action Uncertainty-Driven MCTS Reasoning Mechanism. The comparative results are visualized in Figure~\ref{fig:ablation}.

\paragraph{Impact of Task Uncertainty-Driven Planning}
We analyze the necessity of the adaptive planning mechanism by freezing the agent into static explicit-only or implicit-only modes (blue bars in Figure~\ref{fig:ablation}). The results reveal a distinct domain-dependent preference. On WebArena, the explicit-only mode outperforms the implicit-only mode (39.2\% vs. 38.2\%). The implicit mode struggles to maintain the global thread in deep, structured workflows. Conversely, on WebVoyager, the implicit-only mode surpasses the explicit-only mode (58.1\% vs. 56.6\%). Rigid plans generated by the explicit mode often become obsolete due to high environmental volatility. Crucially, the full WebUncertainty framework consistently achieves the highest performance (63.6\% and 40.1\%). This confirms that task uncertainty effectively signals when to switch between explicit decomposition for stability and reactive stepping for flexibility.

\paragraph{Impact of Action Uncertainty-Driven Reasoning}
The w/o Action Unc. variant (gray bar) removes the ConActU strategy. This reverts the execution phase to standard MCTS and causes the most significant performance degradation. The SR drops by 9.3\% on WebVoyager and 3.4\% on WebArena. The critical flaw of the standard MCTS baseline lies in its inability to decouple error sources. Without EU quantification, the agent cannot identify hallucinations, often wasting search budget expanding nodes on nonexistent elements. Simultaneously, without AU awareness, it treats ambiguous states with multiple valid actions as failures. The agent prunes promising branches instead of triggering necessary exploration. The superior performance of WebUncertainty proves that distinguishing chaotic states from confident failures is essential for robust decision-making.

\subsection{Performance-Cost Analysis}
\label{sec:performance_cost}
MCTS-based reasoning increases computational overhead. However, this is a deliberate trade-off to ensure robustness in complex web tasks. In these scenarios, the cost of a single execution error significantly outweighs the inference cost. 

Importantly, our framework optimizes the MCTS process. It achieves higher performance with lower computational costs than existing search-based methods. As quantified in Table~\ref{tab:efficiency}, we evaluate the average inference time per task on WebVoyager using the Qwen-Max backbone. WebUncertainty reduces the average inference time by over 56\% compared to WebPilot (351.4s vs. 803.7s). It simultaneously improves the SR from 55.8\% to 63.6\%. Future deployments will explore a more systematic performance-cost analysis, including average token usage and total inference cost, to further demonstrate the framework's real-world practicality.
\begin{table}[h]
\centering
\resizebox{0.48\textwidth}{!}{%
\begin{tabular}{lcc}
\toprule
\textbf{Agent} & \textbf{SR (\%)} & \textbf{Avg. Time (s)} \\
\midrule
WebVoyager-rep & 46.5 & 204.9 \\
Browser Use    & 48.8 & 264.6 \\
Agent-E        & 54.3 & 224.7 \\
WebPilot       & 55.8 & 803.7 \\
AgentOccam     & 58.9 & 306.5 \\
\textbf{WebUncertainty} & \textbf{63.6} & \textbf{351.4} \\
\bottomrule
\end{tabular}%
}
\caption{SR and Average Inference Time comparison on WebVoyager using the Qwen-Max backbone.}
\label{tab:efficiency}
\end{table}

\subsection{Sensitivity Analysis}
\label{sec:sensitivity}
To assess the robustness of our framework, we conduct a sensitivity analysis on the planning switch threshold $\delta$ and the evaluation threshold $\tau$. The threshold $\delta$ balances long-horizon coherence from explicit planning with reactive flexibility from implicit planning.

As shown in Table~\ref{tab:sensitivity}, our framework demonstrates strong robustness. It consistently exceeds the strongest baseline (AgentOccam at 58.9\%) across a wide range of values. The framework achieves the optimal SR at $\delta = 0.4$ and $\tau = 6$.

\begin{table}[h]
\centering
\resizebox{0.48\textwidth}{!}{%
\begin{tabular}{lcccccc}
\toprule
\textbf{Threshold} & \multicolumn{6}{c}{\textbf{Values \& SR (\%)}} \\
\midrule
\textbf{$\delta$} & \textbf{0.0} & \textbf{0.2} & \textbf{0.4} & \textbf{0.6} & \textbf{0.8} & \textbf{1.0} \\
\midrule
SR (\%)  & 58.1 & 59.7 & \textbf{63.6} & 62.8 & 58.9 & 56.6 \\
\midrule
\midrule
\textbf{$\tau$} & \textbf{0} & \textbf{2} & \textbf{4} & \textbf{6} & \textbf{8} & \textbf{10} \\
\midrule
SR (\%)  & 52.7 & 57.4 & 62.0 & \textbf{63.6} & 61.2 & 0.0 \\
\bottomrule
\end{tabular}%
}
\caption{Sensitivity analysis of hyperparameters $\delta$ and $\tau$ on WebVoyager (Qwen-Max).}
\label{tab:sensitivity}
\end{table}

\section{Conclusion}
\label{sec:conclusion}

In this work, we presented WebUncertainty, an autonomous agent framework that tackles dynamic interactions and long-horizon execution by modeling dual-level uncertainty. Our Task Uncertainty-Driven Adaptive Planning Mechanism adaptively switches planning modes to ensure robust goal alignment. Furthermore, our Action Uncertainty-Driven MCTS Reasoning Mechanism leverages the ConActU strategy to prune hallucinations and guide decision-making. Extensive experiments on WebArena and WebVoyager demonstrate that WebUncertainty achieves state-of-the-art performance. These results validate the efficacy of integrating uncertainty awareness into the planning and reasoning of web agents.

\section*{Limitations}
Despite its promising performance, WebUncertainty presents several limitations. First, MCTS and multiple candidate generation introduce computational overhead. Although our framework reduces inference time by 56\% compared to WebPilot, this trade-off for robustness may still hinder deployment in real-time or low-cost applications. 

Second, our text-only implementation relies on accessibility trees. The agent may therefore struggle with visually intensive websites where critical information is conveyed through spatial layouts or color coding rather than semantic text. 

Finally, the framework depends on empirical hyperparameters (the thresholds $\delta$ and $\tau$) and the intrinsic calibration of the backbone LLMs. While generally robust, rigid settings may cause suboptimal mode switching in highly volatile environments. Future work will explore adaptive tuning strategies to reduce this dependence.

\section*{Ethics Statement}
This research involves autonomous agents interacting with live web environments. We ensured that all automated interactions were strictly for benign academic purposes, intentionally avoiding malicious actions, unauthorized data collection, or server disruption. Furthermore, as our framework relies on large language models, we acknowledge the inherent risks of propagated biases and hallucinated actions. We strongly advocate for human-in-the-loop oversight before deploying such autonomous agents in critical real-world applications to prevent unintended consequences.

\section*{Acknowledgments}
This paper is funded by National Natural Science Foundation of China (No. 62472138).
\bibliography{custom}

\end{document}